\documentclass{sig-alternate-05-2015}

\begin{document}



\title{Topic Modeling over Short Texts by Incorporating Word Embeddings}


\numberofauthors{8} 
%
\author{
%
%
\alignauthor
Jipeng Qiang\\
       \affaddr{Hefei University of Technology}\\
       \affaddr{Hefei 230009, China}\\
       \email{qjp2100@163.com}
\alignauthor
Ping Chen\\
       \affaddr{University of Massachusetts Boston}\\
       \affaddr{Boston, MA 02155}
\and  
\alignauthor Tong Wang\\
       \affaddr{University of Massachusetts Boston}\\
       \affaddr{Boston, MA 02155}\\
\alignauthor Xindong Wu\\
       \affaddr{University of Louisiana at Lafayette}\\
       \affaddr{Lafayette, Louisiana 70504}\\
}

\maketitle
\begin{abstract}
Inferring topics from the overwhelming amount of short texts becomes a critical but challenging task for many content analysis tasks, such as content charactering, user interest profiling, and emerging topic detecting. Existing methods such as probabilistic latent semantic analysis (PLSA) and latent Dirichlet allocation (LDA) cannot solve this problem very well since only very limited word co-occurrence information is available in short texts. This paper studies how to incorporate the external word correlation knowledge into short texts to improve the coherence of topic modeling. Based on recent results in word embeddings that learn semantically representations for words from a large corpus, we introduce a novel method, Embedding-based Topic Model (ETM), to learn latent topics from short texts. ETM not only solves the problem of very limited word co-occurrence information by aggregating short texts into long pseudo-texts, but also utilizes a Markov Random Field regularized model that gives correlated words a better chance to be put into the same topic. The experiments on real-world datasets validate the effectiveness of our model comparing with the state-of-the-art models.
\end{abstract}



\section{Introduction}

Topic modeling has been proven to be useful for automatic topic discovery from a huge volume of texts. Topic model views texts as a mixture of probabilistic topics, where a topic is represented by a probability distribution over words. Based on the assumption that each text of a collection is modeled over a set of topics, many topic models such as Latent Dirichlet Allocation (LDA) have demonstrated great success on long texts \cite{blei2003latent,griffiths2004finding,wang2015extended}. With the rapid development of the World Wide Web, short text has been an important information source not only in traditional web site, e.g., web page title, text advertisement, and image caption, but in emerging social media, e.g., tweet, status message, and question in Q\&A websites. Compared with long texts, such as news article and academic paper, topic discovery from short texts has the following three challenges: only very limited word co-occurrence information is available, the frequency of words plays a less discriminative role, and the limited contexts make it more difficult to identify the senses of ambiguous words \cite{quan2015short}. Therefore, LDA cannot work very well on short texts \cite{yin2014dirichlet,cheng2014btm}. Finally, how to extract topics from short texts remains a challenging research problem \cite{lau2012line,wang2015exploring}. 

Two major heuristic strategies have been adopted to deal with how to discover the latent topics from short texts. One follows the simple assumption that each text is sampled from only one latent topic which is totally unsuited to long texts, but it can be suitable for short texts compared to the complex assumption that each text is modeled over a set of topics \cite{yan2015probabilistic,zhao2011comparing}. Therefore, many models for short texts were proposed based on this simple assumption \cite{cheng2014btm,yin2014dirichlet}. But, the problem of very limited word co-occurrence information in short texts has not been solved yet. The other strategy takes advantage of various heuristic ties among short texts to aggregate them into long pseudo-texts before topic inference that can help improve word co-occurrence information \cite{mehrotra2013improving,quan2015short,weng2010twitterrank}.  However, these schemes are heuristic and highly dependent on the data, which is not fit for short texts such as news titles, advertisements or image captions. Figure 1 shows an example to explain the shortcomings of existing short text topic models. We can see $s_{1}$ and $s_{2}$ probably include two topics. 'Obama' and 'President' are likely to come from the same topic, and 'NBA' and 'Bulls' are from another topic. The simple assumption that each text is sampled from only one latent topic is unsuited to these texts. And if we directly aggregate the three short texts into two long pseudo-texts, it is very hard to decide how to aggregate these texts since they do not share the same words. But, it is very clear that $s_{1}$ is more similar to $s_{2}$ than $s_{3}$. 

\begin{figure}
	\centering
	\includegraphics[scale=0.4]{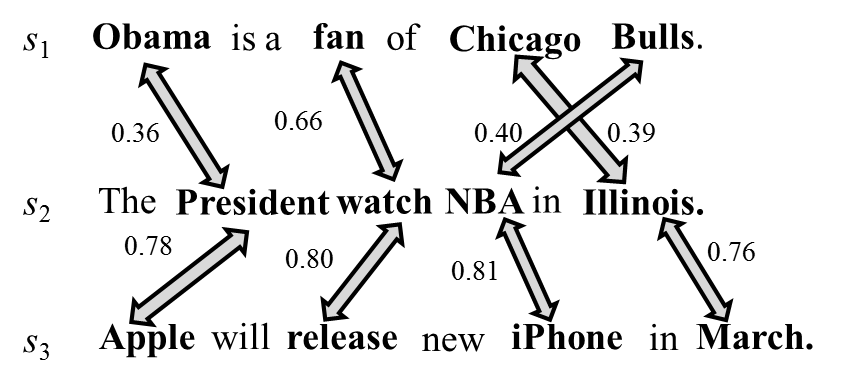}
	\caption{An illustration of the relationship among short texts. There are three short texts, and non-stop words are marked in bold. The shortest distances between two words from different short texts are labeled using the arrows, in which the distance is computed by word embeddings.}
\end{figure}

To overcome these inherent weaknesses and keep the advantages of both strategies, we propose a novel method, Embedding-based Topic Model (ETM), to discover latent topics from short texts. Our method leverages recent results by word embeddings that obtain vector representations for words\cite{mikolov2013distributed,pennington2014glove}. The authors demonstrated that semantic relationships are often preserved in vector operations on word vectors, e.g., vec(\textit{King}) - vec(\textit{man}) + vec(\textit{woman}) is close to vec(\textit{Queen}), where vex(\textit{x}) denotes the vector of word \textit{x}. This suggests that distances between embedded word vectors are to some degree semantically meaningful. For example, all distances in Figure 1 are computed by word embedding model \cite{pennington2014glove}.

ETM has the following three steps. ETM firstly builds distributed word embeddings from a large corpus, and then aggregates short texts into long pseudo-texts by incorporating the semantic knowledge from word embeddings, thus alleviates the problem of very limited word co-occurrence information in short texts. Finally, ETM discovers latent topics from pseudo-texts based on the complex assumption that each text of a collection is modeled over a set of topics. Gaining insights from \cite{xie2015incorporating}, ETM adopts a Markov Random Field regularized model based on collapsed Gibbs sampling which utilizes word embeddings in a soft and topic-dependent manner to improve the coherence of topic modeling. Within a long pseudo-text, if two words are labeled as similar according to word embedding, a binary potential function is defined to encourage them to share the same latent topic. Through this way, ETM can effectively identify the senses of ambiguous words.

To measure the performance of ETM, we conduct experiments on two real-world short text datasets, Tweet 2011 and Google News. Experiments demonstrate that ETM can discover more prominent and coherent topics than the baselines. When applying the learned topic proportions of texts in clustering task, we also find that ETM can infer significantly better topic distribution than the baselines.

The remainder of the paper is organized as follows. Section 2 discusses related work. Section 3 presents the proposed framework for short text topic modeling. Section 4 reports experimental results on real-world datasets.

\section{Related Work}

In this section, we briefly describe the related work from the following two aspects: long text topic modeling and short text topic modeling.

\subsection{Long Text Topic Modeling}

Nigam et al. \cite{nigam2000text} proposed a mixture of unigrams model based on the assumption that each document is generated by one topic. This simple assumption is often too limited to effectively model a large collection of long texts. The complex assumption that each text is modeled over multiple topics was widely used by topic discovery from long texts \cite{hofmann1999probabilistic,blei2003latent,griffiths2004finding}. In a sense, the complex assumption captures the possibility that a document may contain multiple topics. Based on this assumption, many topic models such as Probabilistic Latent Semantic Analysis (PLSA) \cite{hofmann1999probabilistic} and  Latent Dirichlet Allocation (LDA) \cite{blei2003latent} have shown promising results. 

In recent years, knowledge-based topic models have been proposed, which ask human users to provide some prior domain knowledge to guide the model to produce better topics instead of purely relying on how often words co-occur in different contexts. For example, Chen and Liu encode the Must-Links (meaning
that two words should be in the same topic) and Cannot-Links (meaning that two words should not be in the same topic) between words over the topic-word multinomials \cite{chen2014mining}. Besides, two recently proposed models, i.e., a quadratic regularized topic model based on semi-collapsed Gibbs sampler \cite{newman2011improving} and a Markov Random Field regularized Latent Dirichlet Allocation model based on Variational Inference\cite{xie2015incorporating}, share the idea of incorporate the correlation between words. All these models only deal with long texts, and perform poorly on short texts.

\subsection{Short Text Topic Modeling}

The earliest works on short text topic models mainly focused on exploiting external knowledge to enrich the representation of short texts. For instance, Jin et al. \cite{jin2011transferring} first found the related long texts for each short text, and learned topics over short texts and their related long texts using LDA.  Phan et al. \cite{phan2008learning} learned the topics on another large-scale dataset using a conventional topic model such as PLSA and LDA for short text classification. However, these models are only effective when the additional data are closely related to the original data. Furthermore, finding such additional data may be expensive or even impossible.

As a lot of short texts have been collected from social networks such as Twitter, many people analyze this type of data to find latent topics for various tasks, such as event tracking \cite{lin2010pet}, content recommendation \cite{phelan2009using}, and influential users prediction \cite{weng2010twitterrank}. Initially, due to the lack of specific topic models for short texts, some works directly applied long text topic models \cite{ramage2010characterizing,wang2012tm}. Since only very limited word co-occurrence information is available in short texts, some works took advantages of various heuristic ties among short texts to aggregate them into long pseudo-documents before topic inference \cite{mehrotra2013improving,quan2015short}. In a sense, each short text is considered to be generated from a long pseudo-document. The strategy can be regarded as an application of the author-topic model \cite{steyvers2004probabilistic} to tweets, where each tweet (text) has a single author. For example, some models aggregated all the tweets of a user as a pseudo-text \cite{weng2010twitterrank}. As these tweets with the same hashtag may come from a topic, Mehrotra et al. \cite{mehrotra2013improving} aggregated all tweets into a pseudo-text based on hashtags. The other scheme directly aggregates short texts into long pseudo-texts through clustering methods \cite{quan2015short}, in which the clustering method will face this same problem of very limited word co-occurrence information. However, the above approaches cannot be readily applied to more general forms of short texts which provide hardly any such context information.

Recently, some works found that even through the assumption that each text is generated by one topic does not fit long texts, it can work well for short texts \cite{quan2015short,yin2014dirichlet}. Therefore, many topic models adopted this assumption to discover the latent topics in short texts. Zhao et al. \cite{zhao2011comparing} empirically compared the data with traditional news media, and proposed a Twitter-LDA model by assuming that one tweet is generated from one topic. Yin and Wang \cite{yin2014dirichlet} also adopted this assumption for topic inference based on Gibbs sampling. However, these models failed to solve the problem of very limited word co-occurrence information in short texts. Therefore, motivated by the results that prior domain knowledge is useful for long text topic models\cite{newman2011improving,xie2015incorporating}, we will propose a novel method for short texts by incorporating the external word correlation knowledge provided by word embeddings to improve the quality of topic modeling.

\section{Algorithm}

In this section, we discuss our method, Embedding-based Topic Model (ETM), for identifying the key topics underlying a collection of short texts. Our model ETM includes three steps. First, we build distributed word embeddings for the vocabulary of the collection. Second, we aggregate short texts into long pseudo-texts by incorporating the semantic knowledge from word embeddings. We implement K-means using a new metric, Word Mover's Distance (WMD) \cite{kusnerword}, to compute the distance between two short texts.
Third, we adopt a Markov Random Field regularized model which utilizes word embeddings in a soft and topic-dependent manner to improve the coherence of topic modeling. The framework of ETM is shown in Figure 2.

\begin{figure}
	\centering
	\includegraphics[scale=0.5]{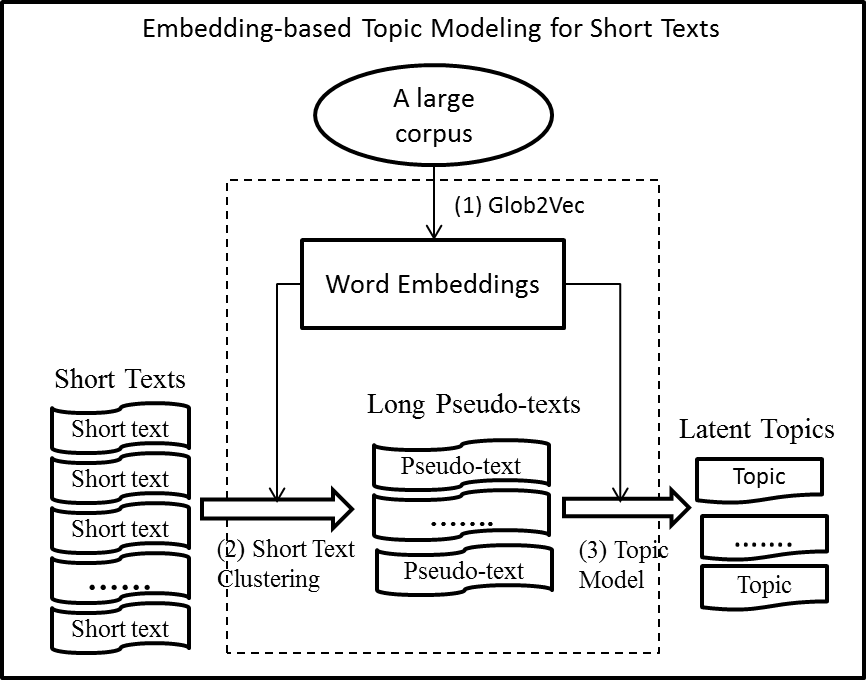}
	\caption{Embedding-based Topic Model for Short Texts}
\end{figure}

\subsection{ Word Embeddings}

Mikolov et al. introduced Word2Vec, to learn a vector representation for each word using a shallow neural network architecture that consists of an input layer, a projection layer, and an output layer to predict nearby words \cite{mikolov2013efficient,mikolov2013distributed}. Word2Vec applies a standard technique such as skip-gram on the given corpus. 
The model avoids non-linear transformations and therefore makes training extremely efficient. This enables learning of embedded word vectors from huge datasets with billions of words, and millions of words in the vocabulary. Word embeddings can capture subtle semantic relationships between words, such as vec(\textit{Berlin}) - vec(\textit{Germany}) + vec(\textit{France}) $\approx$ vec(\textit{Pairs}) and vec(\textit{Einstein}) - vec(\textit{scientist}) + vec(\textit{Picasso}) $\approx$ vec(\textit{painter}), where vec(\textit{x}) denotes the vector of word \textit{x} \cite{mikolov2013distributed}.

Different from Word2Vec that only utilizes local context windows, Pennington et al. later introduced a new global log-bilinear regression model, Glob2Vec, which combines global word-word co-occurrence counts from a corpus, and local context windows based learning similar to Word2Vec to deliver an improved word vector representation. 

\subsection{Short Text Clustering}

After obtaining word embeddings of each word, we use the typical cosine distance measure for the distance between words, i.e., for word vector $v_x$ and word vector $v_y$, we define the distance

\begin{equation} d(v_x, v_y) = 1 - \frac{v_x}{\parallel v_x \parallel _2}\times\frac{v_y}{\parallel v_y \parallel _2}\end{equation}

Consider a collection of short texts, \textit{S} = \{\textit{s}$_{1}$, \textit{s}$_{2}$, ..., \textit{s}$_{i}$, ..., \textit{s}$_{n}$\}, for a vocabulary of \textit{V} words, where \textit{s}$_{i}$ represents the \textit{i}$^{th}$ text. We assume each text is represented as normalized bag-of-words (nBOW) vector,  $\textbf{r}_{i}$ $\in$ $\mathbb{R}$$^{V}$ is the vector of $\textit{s}_{i}$, a \textit{V}-dimension vector, $r_{i,j}$=$\frac{c_{i,j}}{\sum_{v=1}^{V}c_{i,v}}$ where $c_{i,j}$ denotes the occurrence times of the $j^{th}$  word of the vocabulary in text \textit{s}$_{i}$. We can see that a nBOW vector is very sparse as only a few words appear in each text. For example, given three short texts in Figure 1, if we adopt these metrics (e.g.,  Euclidean distance, Manhattan distance \cite{krause2012taxicab}, Cosine Similarity ) to measure distance between two texts, it is hard to find their difference. Therefore, we introduce a new metric, called the \textit{Word Mover's Distance} (WMD)\cite{kusnerword}, to compute the distance between texts. WMD computes the minimum cumulative cost that words from one text need to travel to match exactly the words of the other text as the distance of texts, in which the distance bewteen words is computed by word embeddings.

Let $\textbf{r}_{i}$ and $\textbf{r}_{j}$ be the nBOW representation of $s_i$ and $s_j$. Each word of $\textbf{r}_{i}$ can be allowed to travel to the word of $\textbf{r}_{j}$. Let \textit{T} $\in \mathbb{R}^{m \times m}$ be a flow matrix, where \textit{T}$_{u,v}$ represents how much of the weight of word \textit{u} of $\textbf{r}_{i}$ travels to word \textit{v} of $\textbf{r}_{j}$. To transform all weights of $\textbf{r}_{i}$ into $\textbf{r}_{j}$, we guarantee that the entire outgoing flow from vertex \textit{u} equals to $r_{i,u}$, namely $\sum_{v}^{}T_{u,v} = r_{i,u}$. Correspondingly, the amount of incoming flow to vertex \textit{v} must equal to $r_{j,v}$, namely, $\sum_{u}^{}T_{u,v} = r_{j,v}$. At last, we can define the distance of two texts as the minimum cumulative cost required to flow from all words of one text to the other text, namely,  $\sum_{u,v}^{}T_{u,v} d(u,v)$. Given the constraints, the distance between two texts can be solved using the following linear programming,

\begin{equation}
\begin{split}
& \quad \quad \quad \quad \quad \max_{T \geq 0} \sum_{u,v}^{m}T_{u,v} d(u,v) \\
&such \; that: \sum_{v}^{m}T_{u,v} = r_{i,u} \quad \forall{u} \in \{1,2,...,m\} \\
&\quad \quad \quad \quad \quad \sum_{u}^{m}T_{u,v} = r_{j,v} \quad \forall{v} \in \{1,2,...,m\}
\end{split} 
\end{equation}

The above optimization is a special case of the Earth Mover's Distance (EMD) \cite{rubner1998metric, wolsey2014integer}, a well-known transportation problem for which specialized solvers have been developed \cite{ling2007efficient,pele2009fast}. The best average time complexity of solving the WMD problem is O($m^{3}$log${m}$), where \textit{m} is the number of unique words in the text. To speed up the optimization problem, we relax the WMD optimization problem and remove one of the two constraints. Consequently, the optimization becomes,

\begin{equation}
\begin{split}
& \quad \quad \quad \quad \quad \max_{T \geq 0} \sum_{u,v}^{m}T_{u,v} d(u,v) \\
&such \; that: \sum_{v}^{m}T_{u,v} = r_{i,u} \quad \forall{u} \in \{1,2,...,m\} \\
\end{split} 
\end{equation}

The optimal solution is the probability of each word in one text is moved to the most similar word in the other text. The time complexity of WMD can be reduced to O($m$log${m}$).

Once the distance between texts have been computed, we aggregate short texts into long pseudo-texts based on K-means clustering. Given the number of long pseudo-texts \textit{L}, we compute a score for each short text by averaging the distance between this text and all short texts of each long pseudo-text, that is,
\begin{equation} Score(s_i \in l_j) = \frac{\sum_{s_u\in l_j}d(s_i,s_u)}{\mid l_j \mid}\end{equation}
Where $d(s_i,s_u)$ is the distance between text $s_i$ and $s_u$, $\mid l_j \mid$ represents the number of short texts in pseudo-text $l_j$. In each iteration, for each short text, we choose the smallest score as its long pseudo-text. After a few iterations, we can obtain long pseudo-texts for all short texts.

\subsection{Topic Inference}

In this subsection, we present how to infer the topics from the long pseudo-texts using Markov Random Field Regularized (MRF) Model and parameter estimation based on collapsed Gibbs sampling.

\subsubsection{Model Descritpion}

We adopt the MRF model to learn the latent topics which can incorporate word distances into topic modeling for encouraging words labeled similarly to share the same topic assignment \cite{xie2015incorporating}. Here, we continue to use word embeddings to compute the distance between words. We can see from Figure 3, MRF model extends the standard LDA model \cite{blei2003latent} by imposing a Markov Random Field on the latent topic layer.

Suppose the corpus contains \textit{K} topics and long pseudo-texts with $L$ texts over \textit{V} unique words in the vocabulary. Following the standard LDA, $\Phi$ is represented by a $K \times V$ matrix where the \textit{k}th row $\phi _k$ represents the distribution of words in topic \textit{k}, $\Theta$ is represented by a $L \times K$ where the \textit{l}th row $\theta_l$ represents the topic distribution for the \textit{l}th long pseudo-texts, $\alpha$ and $\beta$ are hyperparameters, $z_{li}$ denotes the topic identities assigned to the $i_{th}$ word in the \textit{l}th long pseudo-text. 

The key idea is that if the distance between two words in one pseudo-text is smaller than a threshold, they are more likely to belong to the same topic. For example, in Figure 1, 'President' and 'Obama' ('Bulls' and 'NBA') are likely to belong to the same topic. Based on this idea, MRF model defines a Markov Random Field over the latent topic. Given a long pseudo-text \textit{l} consisting of $\textit{n}_l$ words $\{{w_{li}}\}^{n_l}_{i=1}$. If the distance between any word pair ($w_{li}$,$w_{lj}$) in \textit{l} is smaller than a threshold, MRF model creates an undirected edge between their topic assignments ($z_{li}$,$z_{lj}$). Finally, MRF creates an undirected graph $\textit{G}_l$ for the \textit{l}th pseudo-text, where nodes are latent topic assignments $\{{z_{li}}\}^{n_l}_{i=1}$ and edges connect the topic assignments of correlated words. For example, in Figure 3, $\textit{G}_l$ is consisted of five nodes ($z_{l1}$,$z_{l2}$,$z_{l3}$,$z_{l4}$,$z_{l5}$) and five edges \{($z_{l1}$,$z_{l2}$,), ($z_{l1}$,$z_{l3}$,), ($z_{l2}$,$z_{l4}$,), ($z_{l2}$,$z_{l5}$,), ($z_{l3}$,$z_{l5}$)\} . 

\begin{figure}
	\centering
	\includegraphics[scale=0.5]{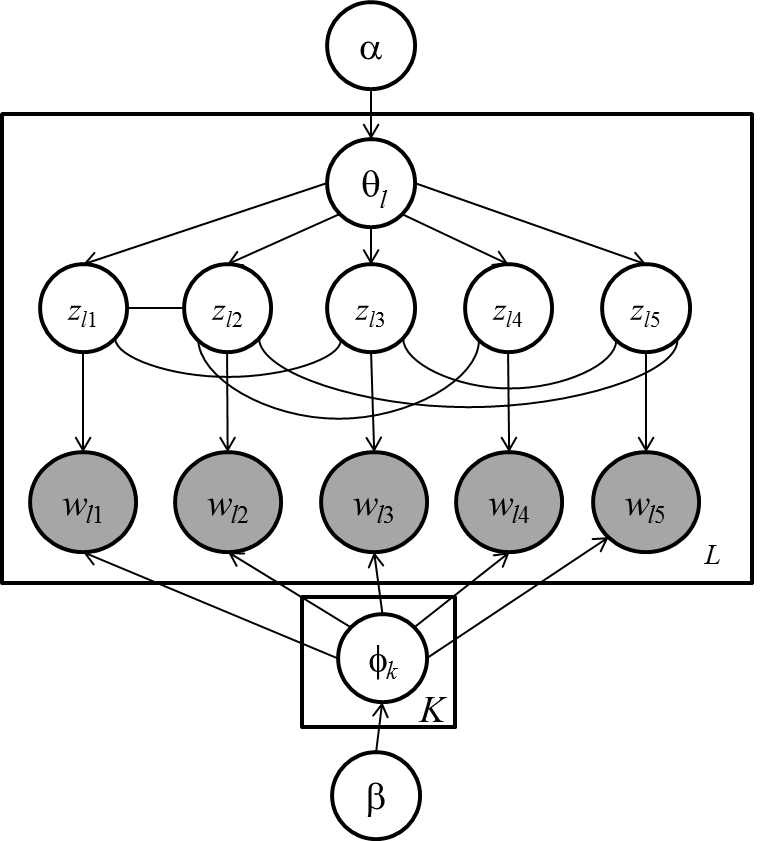}
	\caption{Markov Random Field Regularized Latent Dirichlet Allocation Model}
\end{figure}

The same to LDA, MRF model uses the unary potential for $z_{li}$ as $p(z_{li} \mid \theta_l)$. The difference is MRF model defines binary potential over each edge $(z_{li},z_{lj})$ of $\textit{G}_l$ as $exp\{\mathcal{I}(z_{li}=z_{lj})\}$, which produces a large value if the two topic assignments are the same and generates a small value if the two topic assignments are different, where $\mathcal{I}(\cdot)$ is the indicator function. Hence, similar words in one pseudo-text have a high probability to be put into the same topic. The joint probability of all topic assignments $\textbf{z}_l=\{z_{li}\}_{i=1}^{n_l}$ in MRF model can be calculated as 

\begin{equation}
p(\textbf{z}_l \mid \theta_l, \lambda) =  \prod_{i=1}^{n_l}{p(z_{li} \mid \theta_l)} exp\{\lambda \frac{\sum_{(li,lj)\in \mathcal{P}_l}{\mathcal{I}(z_{li} = z_{lj})}}{\mid \mathcal{P}_l \mid}\}
\end{equation} 
where $\mathcal{P}_l$ represents all edges of $\textit{G}_l$ and ${\mid \mathcal{P}_l \mid}$ is the number of all edges in $\textit{G}_l$. Here, $\lambda$ is a user-specified parameter that controls the tradeoff between unary potential and binary potential. If $\lambda$=0, MRF model is reduced to LDA. Different from LDA that topic label $z_{li}$ is determined by topic distribution $\theta_l$, $z_{li}$ in MRF model depends on both $\theta_l$ and the topic assignments of similar words in the \textit{l}th pseudo-text. 

Formally, the generative process of MRF model is described as follows.

1) Draw $\Theta \sim$ Dirichlet$(\alpha)$ 

2) For each topic $k \in [1, K]$

\hspace{0.6cm} a) draw $\phi_k \sim $ Dirichlet$(\beta)$

3) For each pseudo-text \textit{l} in long pseudo-texts

\hspace{0.6cm} a) draw topic assignments $\textbf{z}_l$ for all words in pseudo-text \textit{l} using Eq.(5)

\hspace{0.6cm} b)	draw $w_{li} \sim $ Multinomial($\phi_{z_{li}}$) for each word in \textit{l}th pseudo-text

There have been a number of inference methods that have been used to estimate the parameters of topic models, from basic expectation maximization \cite{hofmann1999probabilistic}, to approximate inference methods like Variational Inference \cite{blei2003latent} and Gibbs sampling \cite{griffiths2004finding}. Variational Inference tends to approximate some of the parameters, such as $\Phi$ and $\Theta$, not explicitly estimate them, may face the problem of local optimum. Therefore, different from this paper \cite{xie2015incorporating} based on Variational Inference, we will use collapsed Gibbs sampling to estimate parameters under Dirichlet priors in this paper. 

These parameters that need to be estimated include the topic assignments of \textbf{z}, the multinomial distribution parameters $\Phi$ and $\Theta$. Using the technique of collapsed Gibbs sampling, we only need to sample the topic assignments of \textbf{z} by integrating out $\phi$ and $\theta$ according to the following condition distribution:

\begin{equation}
\begin{split}
p(z_{li}=k \mid \textbf{z}_{l,-li}, \textbf{w}_{l,-li})= & (n_{l,-li}^{k}+\alpha)\frac{n_{k,-li}^{w_{li}}+\beta}{n_{k,-li}+V\beta} \\
&exp(\lambda \frac{\sum_{j \in \mathcal{N}_{li}}{(z_{lj}=k)}}{\mid\mathcal{N}_{li}\mid}) 
\end{split}
\end{equation}
where $z_{li}$ denotes the topic assignment for word $w_{li}$ in the \textit{l}th pseudo-text, $\textbf{z}_{l,-li}$ denotes the topic assignments for all words except $w_{li}$ in the \textit{l}th pseudo-text, $n_{l,-li}^{k}$ is the number of times assigned to topic \textit{k} excluding $w_{li}$ in the \textit{l}th pseudo-text, $n_{k,-li}^{w_{li}}$ is the number of times word $w_{li}$ assigned to topic \textit{k} excluding $w_{li}$, $n_{k,-li}$ is the number of occurrences of all words \textit{V} that belongs to topic \textit{k} excluding $w_{li}$, $\mathcal{N}_{li}$ denotes the words that are labeled to be similar to $w_i$ in the \textit{l}th pseudo-text, and $\mid\mathcal{N}_{li}\mid$ is the number of words in $\mathcal{N}_{li}$.

Using the counts of the topic assignments of long pseudo-texts, we can estimate the topic-word distribution of $\phi$ and text-topic distribution $\theta$ as follows,

\begin{equation}
\phi^{w}_{k} = \dfrac{n^w_k+\beta}{n_k+V\beta}
\end{equation}

\begin{equation}
\theta^{l}_{k} = \dfrac{n_{l,k}+\alpha}{n_{l}+K\alpha}  
\end{equation}
where $n^w_k$ is  the number of times word $w$ assigned to topic \textit{k}, $n_{l,k}$ is the number of times word $w_{li}$ assigned to topic \textit{k} in the \textit{l}th pseudo-text, and $n_l$ is the number of words in the \textit{l}th pseudo-text. 

For each short text \textit{s} of long pseudo-texts, we can obtain its topic assignments as follows:

\begin{equation}
p(s = k) = \prod_{j=1}^{n_{s}}{\phi^{w_{j}}_k}
\end{equation}
where $n_s$ is the number of words in short text \textit{s}, $w_j$ is the \textit{j}th word of \textit{s}.

\subsubsection{Parameter Estimation}

There are three types of variables (\textbf{z}, $\Phi$ and $\Theta$) to be estimated for our model ETM. For the \textit{l}th pseudo-text, the joint distribution of all known and hidden variables is given by the hyperparameters:

\begin{equation}
\begin{split}
p(\textbf{z}_l, \theta_l, \textbf{w}_l, \Phi \mid \alpha, \beta, \lambda) = &  p(\Phi | \beta) \cdot \prod_{li=1}^{n_l}{p(w_{li} \mid \phi_{z_{li}})} \\
& \cdot p(\textbf{z}_l \mid \theta_l, \lambda) \cdot p(\theta_l \mid \alpha)
\end{split}
\end{equation}

We can obtain the likelihood of the \textit{l}th pseudo-text $\textbf{w}_l$ of the joint event of all words by integrating out $\phi$ and $\theta$ and summing over $z_{li}$. 

\begin{equation}
p( \textbf{w}_l \mid \alpha, \beta, \lambda) =  \int\int p(\theta_l \mid \alpha) \cdot p(\Phi | \beta) \cdot \prod_{li=1}^{n_l}{p(w_{li} \mid \phi_{z_{li}}, \Phi, \lambda)} 
\end{equation}

Finally, the likelihood of all pseudo-texts $\textbf{W}=\{\textbf{w}_l\}_{l=1}^{L}$ is determined by the product of the likelihood of the independent pseudo-texts:

\begin{equation}
p( \textbf{W} \mid \alpha, \beta, \lambda) =  \prod_{l=1}^{L}{p( \textbf{w}_l \mid \alpha, \beta, \lambda)}
\end{equation}

We try to formally derive the conditional distribution $p(z_{li}=k \mid \textbf{z}_{l,-li}, \textbf{w}_{l,-li})$ used in our ETM algorithm as follows.

\begin{equation}
\begin{split}
p(z_{li}=k \mid \textbf{z}_{l,-li}, \textbf{w}_{l,-li}) &= \frac{p(\textbf{w},\textbf{z} \mid \alpha, \beta, \lambda ) }{p(\textbf{w},\textbf{z}_{l,-li} \mid \alpha, \beta, \lambda)}\\ 
&\propto \frac{p(\textbf{w},\textbf{z} \mid \alpha, \beta, \lambda)}{p(\textbf{w}_{l,-li},\textbf{z}_{l,-li}\mid \alpha, \beta, \lambda)}
\end{split} 
\end{equation}

From the graphical model of ETM, we can see
\begin{equation}
p(\textbf{w},\textbf{z} \mid \alpha, \beta, \lambda ) = p(\textbf{w} \mid \textbf{z}, \beta)p(\textbf{z} \mid \alpha, \lambda) 
\end{equation}

The same to LDA, the target distribution $p(\textbf{w} \mid \textbf{z}, \beta)$ is obtained by integrating over $\phi$, 
\begin{equation}
p(\textbf{w} \mid \textbf{z}, \beta) = \prod_{z_{li}=1}^{K}{\frac{\Delta (\textbf{n}_{z_{li}}+\beta)}{\Delta(\beta)}}, \textbf{n}_{z_{li}} = \{n_{z_{li}}^{(w)}\}_{w=1}^{V}
\end{equation}
where $n_{z_{li}}^{(w)}$ is the number of word \textit{w} occurring in topic $z_{li}$. Here, we adopt the $\Delta$ function in Heinrich (2009), and we can have  $\Delta(\beta)=\frac{\prod^{V}_{w=1}\Gamma(\beta)}{\Gamma(V\beta)}$ and $\Delta (\textbf{n}_{z_{li}}+\beta)=\frac{\prod_{w \in \textbf{w}}\Gamma(n^w_k+\beta)}{\Gamma(n_k+V\beta)}$, where $\Gamma$ denotes the gamma function.

According to Equation (5), we can get
\begin{equation}
p(\textbf{z}_l \mid \theta_l, \lambda) =  exp\{\lambda \frac{\sum_{(li,lj)\in \mathcal{P}_l} \sum_{k=1}^{K}(z_{li} z_{lj})}{\mid \mathcal{P}_l \mid}\} \prod_{k=1}^{K}{\theta_k^{n_l^k}} 
\end{equation}

Similarly, $p(\textbf{z}_l \mid \alpha, \lambda)$ can be obtained by integrating out $\Theta$ as 
\begin{equation}
\begin{split}
p(\textbf{z} \mid \alpha, \lambda) &= \int p(\textbf{z} \mid \Theta, \lambda) p(\Theta \mid \alpha) \\
&= \prod_{l=1}^{L}{exp\{\lambda\frac{\sum_{(li,lj)\in \mathcal{P}_l} \sum_{k=1}^{K}(z_{li} z_{lj})}{\mid\mathcal{P}_l\mid}\}\frac{\Delta (\textbf{n}_l + \alpha)}{\Delta (\alpha)}}
\end{split} 
\end{equation}
where $p(\Theta\mid \alpha)$ is a Dirichlet distribution, and $\textbf{n}_l=\{n_l^{(k)}\}_{k=1}^{K}$.

Finally, we put the joint distribution $p(\textbf{w},\textbf{z} \mid \alpha, \beta, \lambda )$ into Equation (13), the conditional distribution in Equation (6) can be derived 

\begin{equation}
\begin{split}
& p(z_{li}=k \mid \textbf{z}_{l,-li}, \textbf{w}_{l,-li}) \propto \frac{p(\textbf{w},\textbf{z} \mid \alpha, \beta, \lambda)}{p(\textbf{w}_{l,-li},\textbf{z}_{l,-li}\mid \alpha, \beta, \lambda)} \\
& \propto \frac{\Delta (\textbf{n}_l + \alpha)}{\Delta (\textbf{n}_{l, -li} + \alpha)}\frac{\Delta (\textbf{n}_{z_{li}} + \beta)}{\Delta (\textbf{n}_{z_{l,-li}} + \beta)}exp(\lambda \frac{\sum_{j \in \mathcal{N}_{li}}{(z_{lj}=k)}}{\mid\mathcal{N}_{li}\mid}) \\
&\propto (n_{l,-li}^{k}+\alpha)\frac{n_{k,-li}^{w_{li}}+\beta}{n_{k,-li}+V\beta}exp(\lambda \frac{\sum_{j \in \mathcal{N}_{li}}{(z_{lj}=k)}}{\mid\mathcal{N}_{li}\mid})
\end{split} 
\end{equation}

\section{Experiments}

In this section, we show the experimental results to demonstrate the effectiveness of our model by comparing it with five baselines on two real-world datasets.

\subsection{Datasets Description and Setup}

\textbf{Datasets}: We study the empirical performance
of ETM on two short text datasets.

\begin{itemize} 
	\item Tweet2011: Tweet2011 collection is a standard short text collection published on TREC 2011 microblog track\footnote{http://trec.nist.gov/data/tweets/}, which includes approximately 16 million tweets sampled between January 23rd and February 8th, 2011.
	
	\item  GoogleNews: Similar to existing papers \cite{yin2014dirichlet}, we utilize Google news\footnote{http://news.google.com} as a dataset to evaluate the performance of topic models. On Google news dataset, all news articles are grouped into clusters automatically. We took a snapshot of the Google news on April 27, 2015, and crawled the titles of 6,974 news articles belonging to 134 categories. 
\end{itemize}

For each dataset, we conduct the following preprocessing: (1) Convert letters into lowercase; (2) Remove non-latin characters and stop words; (3) Remove words whose length are smaller than 3 or larger than 20; (4) Remove words with frequency less than 3. 

\textbf{Comparison Methods}: We compare our model ETM\footnote{The source code can be downloaded at https://github.com/qiang2100/ETM} with the following baselines:
\begin{itemize} 
	\item Three short text topic models, Unigrams \cite{nigam2000text}, DMM \cite{yin2014dirichlet}, and BTM \cite{cheng2014btm}. Unigrams and DMM use the simple assumption that each text is sampled from only one latent topic. BTM learns topics by directly modeling the generation of word co-occurrence patterns in the corpus.
	\item Two Long text topic models, LDA \cite{griffiths2004finding} and MRF-LDA \cite{xie2015incorporating}. LDA is the most widely used topic model. MRF-LDA is one novel model designed to incorporate word knowledge into topic modeling.  
\end{itemize}

For LDA, we use this package\footnote{http://www.arbylon.net/projects/} and the code\footnote{https://github.com/ariddell/mixture-of-unigrams} for Unigrams, which are provided online. For BTM\footnote{https://github.com/xiaohuiyan/BTM} and MRF-LDA\footnote{http://www.cs.cmu.edu/~pengtaox/}, we use the tools released by the authors. For DMM, we implement its code since the authors did not release the code. 

\textbf{Word Embeddings}: Word2Vec \cite{mikolov2013distributed} and Glob2Vec \cite{pennington2014glove} are different word embeddings. As Glob2Vec has better performance than Word2Vec \cite{pennington2014glove}, the pre-trained embeddings by Glob2Vec based on Wikipedia\footnote{http://nlp.stanford.edu/projects/glove/} is incorporated into our model and MRF-LDA. 

\textbf{Parameter Settings}: For the baselines, we chooses the parameters according to their original papers. For LDA, Unigrams and BTM, both hyperparameters $\alpha$ and $\beta$ are set to 50/\textit{K} and 0.01. For DMM and ETM, both hyperparameters $\alpha$ and $\beta$ are set to 0.1. For MRF-LDA, $\alpha$=0.5 and $\lambda$=1. For ETM, $\lambda$ is set to 1. For our model and MRF-LDA, words pairs with distance lower than 0.4 are labeled as correlated. The number of pseudo-texts is set as \textit{n}/50, where \textit{n} is the number of all short texts in the corpus.  

\subsection{Experimental Results}

\subsubsection{Qualitative Evaluation}

On Tweet2011 dataset, there is no category information for each tweet. Manual labeling might be difficult due to the incomplete and informal content of tweets. Fortunately, some tweets are labeled by their authors with hashtags in the form of '\#keyword' or '@keyword'. We manually choose 10 frequent hashtags as labels and collect documents with their hashtags. These hashtags are 'NBA', 'NASA', 'Art', 'Apple', 'Barackobama', 'Worldprayr', 'Starbucks', 'Job', 'Travel', 'Oscars', respectively. 

\begin{table*}
	\centering
	\caption{Topics learned from Tweets2011 dataset}
	\begin{tabular}{cccc|cccc|cccc}
		\hline
		\multicolumn{4}{c|}{LDA} & \multicolumn{4}{c|}{MRF-LDA}& \multicolumn{4}{c}{Unigrams} \\
		\hline
		Topic 1 & Topic 2 & Topic 3 & Topic 4 & Topic 1 & Topic 2 & Topic 3 & Topic 4 & Topic 1 & Topic 2 & Topic 3 & Topic 4 \\
		(NBA) & (NASA) & (Art) & (Apple) & (NBA) & (NASA) & (Art) & (Apple)& (NBA) & (NASA) & (Art) & (Apple) \\
		\hline
		game & space & artist & apple & game & space & \textbf{time} & iphone & game & space & artist& apple \\
		lebron & shuttle & museum & iphone & lebron & shuttle & artist & apple & lebron & shuttle  & museum & iphone\\
		kobe & launch & \textbf{great} & store & kobe & \textbf{great} & \textbf{video} & \textbf{check} & kobe & launch & \textbf{good} & store \\
		player & nasa & \textbf{check} & \textbf{time} & player & launch & \textbf{twitter} & \textbf{team} & lakers & nasa & artists & \textbf{time} \\
		lakers & atlantis & photo & steve & \textbf{museum} & \textbf{good} & \textbf{blog} & \textbf{live} & player & atlantis & painting& \textbf{good}  \\
		team & \textbf{live} & \textbf{blog} & jobs & lakers & nasa & \textbf{year} & \textbf{love} & team & \textbf{check}& photo & ipod  \\
		coach & video & artists & snow & play & \textbf{store} & \textbf{record} & \textbf{follow} & \textbf{going} & \textbf{live}  & \textbf{blog} & jobs\\
		\textbf{going} & weather & gallery & \textbf{best} & \textbf{tonight} & \textbf{watch} & \textbf{coming} & \textbf{star} &james & \textbf{watch}  & \textbf{check} & video\\
		james & \textbf{watch} & painting & \textbf{good} & james & \textbf{today} & work & \textbf{coach} & play & weather & gallery& snow \\
		points & \textbf{check} & modern & \textbf{google} & \textbf{better} & atlantis & artists & \textbf{going} & allen & crew  & exhibition & steve\\
		
		\hline
		
		 \multicolumn{4}{c}{DMM}  & \multicolumn{4}{c|}{BTM} & \multicolumn{4}{c}{ETM} \\
		\hline
		Topic 1 & Topic 2 & Topic 3 & Topic 4 & Topic 1 & Topic 2 & Topic 3 & Topic 4 & Topic 1 & Topic 2 & Topic 3 & Topic 4 \\
		(NBA) & (NASA) & (Art) & (Apple) &(NBA) & (NASA) & (Art) & (Apple) & (NBA) & (NASA) & (Art) & (Apple) \\
		\hline
		game & space & artist & apple & game & space & artist & apple & game & space & artist & apple\\
		lebron & shuttle & museum & iphone & kobe & shuttle & \textbf{great} & iphone & lebron & shuttle & museum & iphone\\
		kobe & launch & \textbf{check} & store & lebron & launch & museum & \textbf{good} & kobe & launch & writer & store\\
		lakers & nasa & photo & \textbf{time} & lakers & nasa & \textbf{check} & steve & player & nasa & painting & video\\
		player & atlantis & painting & steve & team & atlantis & \textbf{miami} & video & lakers & flight & gallery & ipod\\
		team & \textbf{live} & artists & \textbf{snow} & player & \textbf{live} & painting & store & coach & weather & artists & \textbf{twitter}\\
		james & \textbf{check} & exhibition & jobs & scored & crew & artists & \textbf{time} & points & atlantis & exhibition & tablet\\
		 points & video & modern & \textbf{google} & points & weather & gallery & jobs & james & crew & modern & steve\\
		\textbf{going} & weather & gallery & \textbf{great} & \textbf{going} & \textbf{watch} & \textbf{blog} & ipod & play & image & photo & \textbf{blog}\\
		\textbf{lead} & \textbf{today} & \textbf{blog} & \textbf{good} & james & image & \textbf{free} & \textbf{going} & allen & ares & arts & \textbf{google}\\
		\hline
	\end{tabular}
\end{table*}%

\begin{table*}
	\centering
	\caption{Topics learned from Google news dataset}
	\begin{tabular}{cccc|cccc|cccc}
		\hline
		\multicolumn{4}{c|}{LDA} & \multicolumn{4}{c|}{MRF-LDA}& \multicolumn{4}{c}{Unigrams} \\
		\hline
		Topic 1 & Topic 2 & Topic 3 & Topic 4 & Topic 1 & Topic 2 & Topic 3 & Topic 4 & Topic 1 & Topic 2 & Topic 3 & Topic 4 \\
		(nepal) & (iran) & (bali) & (yemen) & (nepal) & (iran) & (bali) & (yemen) & (nepal) & (iran) & (bali) & (yemen) \\
		\hline
		nepal& iran&bali&yemen&nepal&iran &bali&yemen& nepal& \textbf{nepal}&bali&yemen\\
		death & nuclear&indonesia&\textbf{nepal}&death&meet&indonesia&saudi & quakes& \textbf{quakes}&indonesia&\textbf{iran} \\
		israel& meet&execution&\textbf{toll}&israel&kerry&death&talks & toll& israel&execution&\textbf{nuclear} \\
		quake& kerry&executions&\textbf{death}&quake&nuclear&\textbf{toll}&\textbf{drug} & death& \textbf{rescue}&executions&\textbf{meet} \\
		rescue& zarif&chan&saudi&rescue&zarif&executions&\textbf{iran} & quake& \textbf{quake}&chan&\textbf{kerry}\\
		aid& talks&marries&\textbf{quakes}&aid&\textbf{deal}&execution&strikes & everest& \textbf{aid}&marries&saudi \\
		israeli& \textbf{good}&andrew&\textbf{quake}&everest&\textbf{victim}&chan&yemeni  & \textbf{tops}& israeli&andrew&\textbf{zarif}\\
		israelis& israel&duo&\textbf{iran}&israeli&talks&andrew&war & aid& \textbf{relief}&death&talks\\
		\textbf{relief}& \textbf{foreign}&death&strikes&israelis&powers&marries&saudis & \textbf{rises}& \textbf{help}&duo&arms\\
		\textbf{help}& \textbf{deal}&\textbf{deal}&\textbf{tops}&\textbf{relief}&arms&\textbf{nuclear}&\textbf{babies} & israelis& \textbf{good}&drug&\textbf{israel} \\
		
		\hline
		
		 \multicolumn{4}{c}{DMM}  & \multicolumn{4}{c|}{BTM} & \multicolumn{4}{c}{ETM} \\
		\hline
		Topic 1 & Topic 2 & Topic 3 & Topic 4 & Topic 1 & Topic 2 & Topic 3 & Topic 4 & Topic 1 & Topic 2 & Topic 3 & Topic 4 \\
		(nepal) & (iran) & (bali) & (yemen) & (nepal) & (iran) & (bali) & (yemen) & (nepal) & (iran) & (bali) & (yemen) \\
		\hline
		nepal&iran &bali&yemen & nepal&\textbf{yemen}&bali& \textbf{nepal} &nepal&iran &bali&yemen \\
		israelis&nuclear&indonesia&saudi  & aids&iran&chan& \textbf{bali} &israelis &nuclear&indonesia&saudi \\
		quake&meet&execution&\textbf{iran}  & quake&nuclear&executions& \textbf{chan} &quake &meet&execution&strikes \\
		toll&kerry&executions&strikes & rescue&meet&andrew& \textbf{drug} &toll &kerry&executions&yemeni \\
		israel&zarif&chan&yemeni  & israel&\textbf{saudi}&marries& \textbf{aids} &israel &zarif&chan&saudis \\
		rescue&israel&marries&saudis & toll&\textbf{kerry}&sukumaran& \textbf{arms}  &rescue &israel&marries&strikes \\
		death&weapon&andrew&talks & death&arms&\textbf{final}& \textbf{claims} &death &weapon&andrew&war \\
		aid&npt&death&strike & aid&talks&duo& \textbf{death} &aid &npt&death&talks \\
		everest&\textbf{deal}&duo&\textbf{tops} & everest&zarif&myuran& \textbf{duo} &everest &talks&duo&houthis \\
		israeli&\textbf{foreign}&durg&houthis & israeli&\textbf{good}&\textbf{ahead}& \textbf{pair} &israeli &powers&durg&arms \\
		\hline
	\end{tabular}
\end{table*}%

Table 1 shows some topics learned by the six models on the Tweet2011 dataset. Each topic is visualized by the top ten words. Words that are noisy and lack of representativeness are highlighted in bold. These four topics are about 'NBA', 'NASA', 'Art' and 'Apple', respectively. From Tabel 1, our model ETM can learn more coherent topics with fewer noisy and meaningless words than all baseline models. Long text topic models (LDA and MRF-LDA) that model each text as a mixture of topics does not fit for short texts, as short text suffers from the sparsity of word co-occurrence patterns. Because short text only consists of a few words, MRF-LDA incorporating word correlation knowledge cannot improve the coherence of topic modeling. Consequently, noise words such as \textit{better}, \textit{great}, \textit{good}, \textit{today} which cannot effectively represent a topic due to their high frequency. Compared to long text, short text probably contains only one topic. Therefore, short text topic models (Unigrams and DMM) adopt a simple assumption that each text is generated by one topic work well on short texts compared to long text topic models. Similar to Unigrams and DMM, BTM posits that unordered word-pair co-occurring in a short text share the same topic drawn from a mixture of topics that can help improve the coherence of topic modeling. But, the existing short text topic models suffer from two problems. On one hand, the frequency of words in short text plays a less discriminative role than long text, making it hard to infer which words are more correlated in each text. On the other hand, these models bring in little additional word co-occurrence information and cannot alleviate the sparsity problem. As a consequence, the topics extracted from these three short text topic models are not satisfying. For example, Topic 3 learned by Unigrams contains less relevant words such as \textit{blog}, \textit{good}, and \textit{check}.  The Apple topic (Topic 4) by DMM and BTM consists of meaning-less words such as \textit{time}, \textit{good}, etc. 

Our method ETM incorporates the word correlation knowledge provided by words embedding over the latent topic to cluster short texts to generate long pseudo-text. In this condition, the frequency of words in pseudo-text plays an important role to discover the topics based on this assumption each text is modeled as a mixture of topics. Simultaneously, our model ETM uses the word correlation knowledge over the latent topic to encourage correlated words to share the same topic label. Hence, although similar words may not have  high co-occurrence in the corpus, they remain have a high probability to be put into the same topic. Consequently, from Table 1 we can see that the topics learned by our model are far better than those learned by the baselines. The learned topics have high coherence and contain fewer noisy and irrelevant words. Our model also can recognize the topic words that only have a few occurrences in the collection. For instance, the word \textit{flight} from Topic 2, \textit{writer} from topic 3, and \textit{tablet} can only be recognized by our model ETM. 

Table 2 shows some topics leaned from GoogleNews dataset. The four topics are events on April 27, 2015, which are "Nepal earthquake", "Iran nuclear", "Indonesia Bali", and "Yemen airstrikes". From this table, we observe that the topic learned by our method are not only better in coherence than those learned from long text topic models (LDA and MRF-LDA), but better than short text topic models (Unigrams, DMM, and BTM), which again demonstrates the effectiveness of our model. 

\subsubsection{Quantitative Evaluation}
Similar to \cite{xie2013integrating,xie2015incorporating}, we also evaluate our model in a quantitative manner based on the coherence measure (CM) to assess how coherent the learned topics are. For each topic, we choose the top 10 candidate words and ask human annotators to judge whether they are relevant to the corresponding topic. First, annotators need to judge whether a topic is interpretable or not. If not, the 10 words of the topic are labeled as irrelevant, or which words are identified by annotators as relevant words for this topic. Coherence measure (CM) is defined as the ratio of the number of relevant words to the total number of candidate words. In our experiments, four graduate students participated the labeling. For each dataset, we choose 10 topics for labeling. 

\begin{table}
	\centering
	\caption{CM (\%) on Tweet2011 Dataset}
	\begin{tabular}{|c|c|c|c|c|c|}
		\hline
		Method & A1 &  A2 & A3 & A4 & Mean\\
		\hline
		LDA & 54 & 42 & 45 & 67 & 52 $\pm$ 11.2 \\
		\hline
		MRF-LDA & 14 & 16 & 16 & 27 & 18.2 $\pm$  5.9 \\
		\hline
		Unigrams & 66  & 45 & 56 & 59 & 56.5  $\pm$ 8.7 \\
		\hline
		DMM & 70 & 49 & 50 & 60 & 57.2 $\pm$  9.8\\
		\hline
		BTM & 62 & 45 & 50 & 77 & 58.5  $\pm$  14.2\\
		\hline
		ETM & \textbf{72} & \textbf{62} & \textbf{73} & \textbf{83} & \textbf{72.5} $\pm$   \textbf{8.5}\\
		\hline
	\end{tabular}
\end{table}%

Table 3 and Table 4 show the coherence measure of topics inferred on Tweet2011 and GoogleNews datasets, respectively. We can see our model ETM significantly outperforms the baseline models. On Tweet2011 dataset, ETM achieves an average coherence measure of 72.5\%, which is larger than long text topic models (LDA and MRF-LDA) with a large margin. Compared to short text topic models, ETM still has a big improvement. In GoogleNews dataset, our model is also much better than the baselines. In conclusion, our model produces better results on both datasets compared to the baselines, which demonstrate the effectiveness of our model in exploring word correlation knowledge from words embeddings to improve the quality of topic modeling.

\begin{table}
	\centering
	\caption{CM (\%) on GoogleNews Dataset}
	\begin{tabular}{|c|c|c|c|c|c|}
		\hline
		Method & A1 &  A2 & A3 & A4 & Mean \\
		\hline
		LDA & 95 & 95 & 79& 95& 91 $\pm$  8\\
		\hline
		MRF-LDA & 81&75 & 70&64 & 72.5 $\pm$ 7.2\\
		\hline
		Unigrams & 88 & 73& 79& 94& 83.5 $\pm$  9.3\\
		\hline
		DMM & 94 &93 & 90 & 93& 92.5 $\pm$  1.7\\
		\hline
		BTM & 80& 85 &75 & 78& 79.5 $\pm$ 4.2\\
		\hline
		ETM & \textbf{96}& \textbf{96} & \textbf{94}&\textbf{96} & \textbf{95.5} $\pm$  \textbf{1.0}\\
		\hline
	\end{tabular}
\end{table}%

\subsubsection{Short Text Clustering}

We further compare the performance of all models in clustering on Tweet2011 and GoogleNews datasets. To provide alternative metrics, the normalized mutual information (NMI) is used to evaluate the quality of a clustering solution \cite{huang2013dirichlet,yin2014dirichlet}. NMI is an external clustering validation metric that effectively measures the amount of statistical information shared by the random variables representing the cluster assignments and the user-labeled class assignments of the data points. NMI value is always a number between 0 and 1, where 1 represents the best result and 0 means a random text partitioning. We run each model 20 times on each dataset and report the mean and standard deviation of their NMI values. 

Table 5 shows the performance of all models on the two datasets. First, we can see that ETM performs significantly better than long text topic models (LDA and MRF-LDA). This is because long text topic models do not consider the problem that only very limited word co-occurrence information is available in short texts. Second, we can find that ETM outperforms short text topic models (Unigrams, DMM, and BTM). This is because topic models lack the mechanism to incorporate word correlation knowledge and generate the words independently. Therefore, we can conclude that our model fits for short texts compared to the baselines. Meanwhile, from the standard deviation of all results, we can see that the standard deviation of ETM is smaller than all other methods. This demonstrates that our assumption for short texts can reasonably simulate the generative process.

\begin{table}
	\centering
	\caption{NMI values on Tweet2011 and GoogleNews }
	\begin{tabular}{|c|c|c|}
		\hline
		Method & Tweet2011 &  GoogleNews   \\
		\hline
		LDA & 0.2809 $\pm$ 0.0037 &  0.8669 $\pm$ 0.0145  \\
		\hline
		MRF-LDA & 0.0504 $\pm$ 0.0119 & 0.6642 $\pm$ 0.0354 \\
		\hline
		Unigrams & 0.3250 $\pm$ 0.0139& 0.8948  $\pm$ 0.0184 \\
		\hline
		DMM & 0.3151$\pm$  0.0126& 0.8963 $\pm$ 0.0501\\
		\hline
		BTM & 0.0624 $\pm$  0.0123 & 0.8885 $\pm$  0.0319\\
		\hline
		ETM & \textbf{0.3999} $\pm$ \textbf{0.0095} & \textbf{0.9193} $\pm$ \textbf{0.0193} \\
		\hline
	\end{tabular}
\end{table}%

\section{Conclusion}

We propose a novel model, Embedding-based Topic Modeling (ETM), to discover the latent topics from short texts. ETM first aggregates short texts into long pseudo-texts by incorporating the semantic knowledge from word embeddings, then infers topics from long pseudo-texts using Markov Random Field regularized model, which encourages words labeled as similar to share the same topic assignment. Therefore, by incorporating the semantic knowledge ETM can alleviate the problem of very limited word co-occurrence information in short texts . Experimental results on two real-world short datasets corroborate its effectiveness both qualitatively and quantitatively over the state-of-the-art methods.




\bibliographystyle{abbrv}
\bibliographystyle{abbrv}
\bibliography{sigproc}  







\end{document}